\documentclass{article}

\usepackage[preprint]{nips_2018}

\usepackage[utf8]{inputenc} 
\usepackage[T1]{fontenc}    
\usepackage{hyperref}       
\usepackage{url}            
\usepackage{booktabs}       
\usepackage{amsfonts}       
\usepackage{nicefrac}       
\usepackage{microtype}      
\usepackage{bm}
\usepackage{hyperref}
\usepackage{natbib}
\usepackage{amsmath}
\usepackage{amssymb}
\usepackage{graphicx}
\usepackage{subfigure}
\usepackage{algorithm}
\usepackage{algorithmic}
\usepackage{color}

\setcitestyle{numbers,square,comma,sort&compress}

\title{Multi-Statistic Approximate Bayesian Computation with Multi-Armed Bandits}

\author{
  Prashant Singh\\
  Division of Scientific Computing\\
  Department of Information Technology\\
  Uppsala University \\
  SE-751 05, Uppsala \\
  Sweden\\
  \texttt{prashant.singh@it.uu.se} \\
  \And
  Andreas Hellander\\
  Division of Scientific Computing\\
  Department of Information Technology\\
  Uppsala University \\
  SE-751 05, Uppsala \\
  Sweden\\
  \texttt{andreas.hellander@it.uu.se} \\
}

\begin{document}

\maketitle

\begin{abstract}
  Approximate Bayesian computation is an established and popular method for likelihood-free inference with applications in many disciplines. The effectiveness of the method depends critically on the availability of well performing summary statistics. Summary statistic selection relies heavily on domain knowledge and carefully engineered features, and can be a laborious time consuming process. Since the method is sensitive to data dimensionality, the process of selecting summary statistics must balance the need to include informative statistics and the dimensionality of the feature vector. This paper proposes to treat the problem of dynamically selecting an appropriate summary statistic from a given pool of candidate summary statistics as a multi-armed bandit problem. This allows approximate Bayesian computation rejection sampling to dynamically focus on a distribution over well performing summary statistics as opposed to a fixed set of statistics.
The proposed method is unique in that it does not require any pre-processing and is scalable to a large number of candidate statistics. This enables efficient use of a large library of possible time series summary statistics without prior feature engineering. 
The proposed approach is compared to state-of-the-art methods for summary statistics selection using a challenging test problem from the systems biology literature.
\end{abstract}

\section{Introduction}
\label{sec:intro}
The use of modeling and simulation techniques to supplement real world experiments and observations often involves fitting the parameters of a stochastic simulator or model to observed data. Once the simulator has been tuned to agree with observed data, it can be used to study the corresponding natural process and to generate hypotheses. This \emph{parameter inference} problem is routinely encountered in various fields of science and engineering \cite{beck1977parameter, ashyraliyev2009systems, cash1979parameter}. 

Likelihood-based inference \cite{severini2000likelihood} is an intuitive approach that involves deriving a likelihood function that describes the probability of the data, given the parameters. However, due to the complexity of simulators and models involved, it is often not possible to derive an analytical form of the likelihood function. For stochastic models, such as models of gene regulation in systems biology \cite{Elowitz2002-ho}, the likelihood function can be formulated but the cost of computing it is prohibitive. In such situations, approximate Bayesian computation (ABC) \cite{sunnaaker2013approximate} is a popular methodology for likelihood-free approximate inference.  ABC-based approaches solve the inference problem by computing the deviation between simulated data and the observed data, and accepting or rejecting candidate parameters based on the deviation as measured by a chosen distance function.  
In the simplest form, this process is  repeated until $N$ samples have been accepted, where $N$ is a user-defined constant. The inferred parameters are then reported as the mean of the parameters corresponding to accepted samples.

The probability of generating samples that are accepted diminishes with increase in data dimensionality \cite{sunnaaker2013approximate}. A common way of alleviating this problem is by making use of \emph{summary statistics} that efficiently capture the important information and patterns within the simulated result (e.g., a lengthy time series) as a low-dimensional representation (e.g., temporal mean). The distance function then operates upon the summary statistic(s) instead of the complete simulation results. Popular summary statistics include simple statistics such as mean and higher-order moments, median, mode and frequency, but can also be more complex and be based on problem specific knowledge.  

The choice of a summary statistic, given a black-box parameter inference problem is not straightforward. While a large pool of automatically extracted statistics associated with time series analysis \cite{Fulcher2013-yw} increases the chance that informative statistics are included, the variance of the estimator increases rapidly with the number of used statistics, degrading the performance of ABC. However, using a small number of chosen statistics increases the risk of poor performance for the specific problem. In absence of problem-specific information, a concrete and principled approach is needed to automatically select the most applicable summary statistic for parameter inference from a larger pool (tens to hundreds) of candidate statistics. This paper formulates the problem of dynamically finding the most informative summary statistic in each iteration of the parameter inference process as a multi-armed bandit (MAB) problem \cite{vermorel2005multi}, wherein selection of the appropriate statistic corresponds to pulling the optimal arm in a bandit setting in each iteration of the algorithm, thereby maximizing the reward or minimizing the regret (minimizing the cumulative distance function value corresponding to the sequence of arm pulls). To the best of the authors' knowledge, this is the first approach exploring summary statistic selection in a MAB context. In contrast to many popular methods for summary statistics selection \cite{prangle2015summary}, the proposed method is computationally inexpensive and does not require any \emph{a priori} pre-processing.

The paper is organized as follows. Section \ref{sec:abc} introduces ABC, discusses the significance of summary statistics and surveys related work. Section \ref{sec:mab} concretely defines the problem and describes the novel MAB-inspired approach towards parameter inference. Section \ref{sec:experiments} demonstrates the efficiency and robustness of the proposed method by inferring parameters of a complex stochastic biochemical reaction network from systems biology. Section \ref{sec:discussion} presents a detailed analysis of the proposed method and identifies future research directions. Section \ref{sec:conclusion} concludes the paper. 

\section{Approximate Bayesian computation and summary statistic selection}
\label{sec:abc}
Let $f\colon({\bm \theta}, V)\to {\bf y}$ be a function representing the simulation-based model that maps model parameters ${\bm \theta}$ and certain random variables $V$ to responses ${\bf y}$. The random variables $V$ are part of the simulator and represent the inherent stochasticity of the  process, while ${\bm \theta}$ represents the control parameters of the process. For a fixed set of parameters ${\bm \theta}$, the responses ${\bf y}$ fluctuate randomly over multiple simulations owing to the stochasticity represented by $V$.

Due to the presence of $V$, it is implicitly possible to define a random variable ${\bf y}_{{\bm \theta}}$ corresponding to the simulation model \cite{diggle1984monte,lintusaari2017fundamentals}. For observed data ${\bf y}_o$, and a given parameter combination ${\bm \theta}$, the probability of ${\bf y}_{{\bm \theta}}$ assuming a value in a $\tau$-neighborhood $B_{\tau}({\bf y}_o)$ around ${\bf y}_o$ equals the probability of  sampling values of $V$ lying in the neighborhood $P({\bf y}_{{\bm \theta}} \in B_{\tau}({\bf y}_o)) = P(f({\bm \theta}, V) \in B_{\tau}({\bf y}_o))$ \cite{lintusaari2017fundamentals}.
Let $L({\bm \theta})$ be the likelihood corresponding to parameters ${\bm \theta}$, then,
\begin{equation}
\label{eq:origLikelihood}
L({{\bm \theta}}) \propto P({\bf y}_{{\bm \theta}} \in B_{\tau}({\bf y}_o)).
\end{equation}
As the size of the neighborhood $\tau$ approaches 0, $L({\bm \theta}) = \lim_{\tau \rightarrow 0} c_{\tau} P({\bf y}_{{\bm \theta}} \in B_{\tau}({\bf y}_o)),$
where $c_{\tau}$ is a constant of proportionality that depends on $\tau$  \cite{lintusaari2017fundamentals}. In case of the random variable ${\bf y}_{{\bm \theta}}$ being discrete, $L({\bm \theta}) = P({\bf y}_{{\bm \theta}} = {\bf y}_o)$.

Let $p({{\bm \theta}})$ be the prior distribution of parameters ${\bm \theta}$. The posterior distribution $p({\bm \theta} \mid {\bf y}_o)$ can be estimated by accepting samples with probability proportional to $L({\bm \theta})$. The accepted samples correspond to simulated output being equal to a sample from the fixed dataset.

However, for complex problems encountered in the real world, $P({\bf y}_{{\bm \theta}} = {\bf y}_o)$ is very often exceedingly small. Therefore, the acceptance condition is relaxed to accept samples within a distance $\tau$, as computed according a chosen distance function $d$ as,
\begin{equation}
\label{eq:dist}
d({\bf y}_{{\bm \theta}}, {\bf y}_o) \leq \tau.
\end{equation}
This forms the basis on which ABC techniques estimate the likelihood of the fixed dataset, given the parameter values. The popular ABC rejection sampling method \cite{pritchard1999population,lintusaari2017fundamentals} involves iteratively sampling and simulating candidate parameter combinations ${\bm \theta} \sim p({\bm \theta})$, with ${\bm \theta}$ being accepted according to Eq. \ref{eq:dist}. The accepted samples form the set ${\bm \theta}_{acc}$.

The accepted samples ${\bm \theta}_{acc}$ do not belong to the posterior $p({\bm \theta} \mid {\bf y}_o)$, but to an approximation $p_{d,\tau}({\bm \theta} \mid {\bf y}_o)$ due to the relaxed condition (Eq. \ref{eq:dist}). The likelihood corresponding to Eq. \ref{eq:origLikelihood} becomes $L_{d,\tau}({\bm \theta}) \propto P(d({\bf y}_{{\bm \theta}}, {\bf y}_o) \leq \tau)$,
leading to the posterior, $p_{d,\tau}({\bm \theta} \mid {\bf y}_o) \propto P(d({\bf y}_{{\bm \theta}}, {\bf y}_o) \leq \tau) p({\bm \theta}).$

The distance function $d$ is a crucial part of the approximation, and consequently, of the parameter inference process. The fixed data ${\bf y}_o$ and simulated responses ${\bf y}_{{\bm \theta}}$ are reduced to one or more summary statistics or high-level features, that capture certain types of behavior present within data. Popular summary statistics include mean, variance, entropy, autocorrelation, etc. Equation \ref{eq:dist} is modified to incorporate the use of summary statistic $s$ as $d(s({\bf y}_{{\bm \theta}}), s({\bf y}_o)) \leq \tau.$

The use of summary statistics is also a part of the approximation, in addition to the relaxed distance computation in Eq. \ref{eq:dist}. By reducing the complete datasets to a handful of features, there is a risk of losing important information relevant to effective parameter inference \cite{csillery2010approximate}. However, using too many summary statistics may lead to large approximation errors due to the `curse of dimensionality' \cite{beaumont2002approximate,prangle2015summary}. Indeed, \citet{barber2015rate} studied the mean squared error (MSE) of a Monte Carlo estimate obtained by the rejection sampling algorithm. They showed that with optimal hyperparameter tuning of the ABC rejection sampler and under the considered regularity conditions, MSE was $\mathcal{O}(n^{-4/(q+4)})$ with $q$ being the number of summary statistics used, and $n$ being the number of simulated datasets. The result holds for large $n$ and $\tau$ close to 0. As a consequence, selection of appropriate summary statistics is an extremely crucial task for high-quality parameter inference \cite{fearnhead2012constructing,nunes2010optimal,barnes2012considerate,prangle2014semi}.
\citet{prangle2015summary} divides methods into the categories of subset selection, projection, and auxiliary likelihood. The following subsections briefly review each of these categories.


\subsection{Subset selection} 
In subset selection methods, a subset of $S$ statistics is selected, that typically optimizes some criterion on training data. The training data is simulated beforehand. The method of approximate sufficiency \cite{joyce2008approximately} adds/removes a candidate summary statistics to/from a subset, and measures the resulting effect on the ABC posterior. This requires approximating the posterior using ABC.
The authors note that implementation of the method is not obvious in higher dimensional parameter spaces \cite{joyce2008approximately}. \citet{nunes2010optimal} propose a method to find a $S$ that minimizes the entropy of the resulting ABC posterior. Entropy is used as a measure of informativeness of the posterior, with lower entropy being more informative. \citet{blum2013comparative} argue that lower values of entropy may not always correspond to more accurate inference. \citet{barnes2012considerate} define sufficient statistics as summary statistics that maximize mutual information between $S'({\bf y_{\bf \theta}})$ and ${\bm \theta}$. They derive an expression for necessary conditions for sufficiency of $S'({\bf y_{\bf \theta}})$ as the Kullbeck-Leibler (KL) divergence of $P({\bm \theta} \mid S'({\bf y}_o))$ from $P({\bm \theta} \mid {\bf y}_o)$ being $0$ \cite{barnes2012considerate,prangle2015summary}. This provides a test for adding a candidate statistic to an existing subset of statistics $S_1({\bf y_{\bf \theta}})$ to obtain $S_2({\bf y_{\bf \theta}})$. $S_2$ is more informative than $S_1$ if the estimated KL divergence of $P_{ABC}({\bm \theta} \mid S_1({\bf y}_o))$ from $P_{ABC}({\bm \theta} \mid S_2({\bf y}_o))$ is above a specified threshold. Setting the threshold, and finding a sufficient subset requires multiple ABC runs \cite{barnes2012considerate,prangle2015summary}. 

All approaches mentioned above require multiple ABC runs as pre-processing in order to obtain an informative subset of summary statistics from a larger candidate statistics pool. Regularization approaches \cite{blum2010choosing,sedki2012contribution,blum2013comparative} do not require ABC runs, but require a training set. A linear regression estimator is trained that maps covariates $S$ to responses ${\bm \theta}$. Variable selection is then performed to find an informative subset. In summary, subset selection methods are interpretable, but can be extremely computationally expensive \cite{prangle2015summary}. For a deeper discussion, the reader is referred to \citet{prangle2015summary}.

\subsection{Projection}
Projection methods start with a set of candidate summary statistics $S$, and aim to find an informative lower dimensional projection of $S$, e.g., a linear transformation. This requires a training set of simulated values, and application of dimensionality reduction techniques such as partial least squares \cite{wegmann2009efficient}, linear regression \cite{fearnhead2012constructing} and boosting \cite{aeschbacher2012novel}. Partial least squares is well-known, but lacks theoretical support for use in ABC and is reported to perform poorly \cite{blum2013comparative}. Linear regression and boosting have been shown to perform well \cite{prangle2015summary}. Projection methods are computationally cheaper than subset selection methods as they avoid repeated subset calculations. They also explore a larger space of summaries. However, as new statistics lie in a space different from original summary statistics, projection methods are less interpretable than subset selection methods.

\subsection{Auxiliary likelihoods}
Auxiliary likelihood-based methods use performance of summary statistics on simpler, similar problems as an indicator of which statistics are likely to perform well for a complex problem at hand. Essentially, a simpler and tractable likelihood is specified, and used to derive summary statistics \cite{pritchard1999population,fu1997estimating}. Auxiliary likelihood-based methods rephrase the search for informative summary statistics, as a search for informative auxiliary likelihoods. Unlike subset selection and projection methods, the need for multiple ABC runs as pre-processing or training data can be eliminated if domain knowledge of similar, simpler problems is available. However, in absence of such knowledge, as is often the case when using black-box simulators, training data will be needed to construct auxiliary likelihoods. A deeper discussion can be found in \citet{prangle2015summary}.

\section{Dynamic summary statistics selection using multi-armed bandits}
\label{sec:mab}



Let ${\bm {\bm \theta}}$ represent the parameters of the simulator, and let $S = (s_1, s_2, ..., s_K)$ be a pool of $K$ candidate summary statistics. Consider a given parameter inference problem to be solved using the ABC rejection sampling algorithm. The problem is described by the simulator $f$, the observed data set ${\bf y}_o$, the prior $p(\cdot)$, the distance function $d$, the desired number of accepted samples $N$, the acceptance threshold $\tau$ and a pool of candidate summary statistics $S$. Algorithm \ref{alg:rejectionStat} outlines the proposed dynamic approach. The difference from standard ABC rejection sampling is the use of a method, $SelectStatistic$, that in each iteration identifies the summary statistic $s$ to use from the global pool $S$, with the aim of minimizing the number of simulations needed to attain $N$ accepted samples forming ${\bm \theta}_{acc}$. Here,  $SelectStatistic$ may make use of past values of the distances calculated for each summary statistic. 

It should be noted that dynamic selection (and the resulting variability) of the selected statistic between rejection sampling iterations has consequences. The traditional rejection sampler identifies a distribution over the static set of supplied summary statistics, while Algorithm \ref{alg:rejectionStat} identifies a distribution over well performing summary statistics. The traditional rejection sampler does not incorporate summary statistic selection, while Algorithm \ref{alg:rejectionStat} performs on-the-fly selection. Towards this end, the distances corresponding to all candidate summary statistics are normalized in $[0,1]$.



\begin{algorithm}
   \caption{The ABC rejection sampling algorithm with dynamic summary statistic selection.}
   \label{alg:rejectionStat}
\begin{algorithmic}[1]
   \STATE {\bfseries Input:} simulator $f$, observed data ${\bf y}_o$, prior $p(\cdot)$, distance function $d$, accepted samples count $N$, threshold $\tau$, summary statistic pool $S$
   \STATE {\bfseries Output:} accepted samples ${\bm \theta}_{acc}$
   \FOR{$i=1$ {\bfseries to} $N$}
   		\REPEAT
        	\STATE ${\bm \theta} \sim p({\bm \theta})$
            \STATE ${\bf y}_{{\bm \theta}} \gets f({\bm \theta})$
            \STATE $s \gets SelectStatistic(S)$
        \UNTIL{$d(s({\bf y}_{{\bm \theta}}), s({\bf y}_o)) \leq \tau$}
        \STATE ${{\bm \theta}}_{acc}^{(i)} \gets {\bm \theta}$
   \ENDFOR
\end{algorithmic}
\end{algorithm}



The $SelectStatistic$ method is constructed here by solving a multi-armed bandit (MAB) problem. The MAB problem deals with the trade-off that an agent faces between exploring the given environment to obtain new information, and exploiting existing knowledge to select the future course of action. The problem was introduced by \citet{robbins1985some}, and has extensively been used since in a variety of applications \cite{kuleshov2014algorithms}, such as clinical trials, engineering design, recommender systems, etc.

A MAB problem derives its name from the problem setting of a gambler playing a slot machine in a casino. The slot machine consists of multiple arms, and the gambler aims to maximize the amount of money he collects in successive arm pulls of the machine. 
In the stochastic formulation \cite{vermorel2005multi}, the problem consists of $K$ probability distributions $(P_1,...,P_K)$ corresponding to $K$ arms, with associated means $(\mu_1,...,\mu_K)$ and variances $(\sigma_1^2,...,\sigma_K^2)$. Each instance of an arm pull $m$ results in receiving a reward $r_i^m$, where $i$ is the index of the pulled arm.
The probability distributions are initially unknown, and the goal is to infer the distribution with the highest expected value, while also maximizing the rewards \cite{vermorel2005multi}. 

Let $S = \{s_1, s_2, ..., s_K\}$ be the user-supplied set of $K \geq 2$ summary statistics. Each summary statistic $s_i$ corresponds to an arm in the multi-armed bandit setting, with probability distribution $P_i$ and corresponding mean $\mu_i$. Selection of, or pulling the arm $s_i$ is associated with reward $r_i$ sampled from $P_i$. 
We assume $r_i = -d({s_i(\bf y}_{\bm \theta}), s_i({\bf y}_{o}))$, where $d$ is a chosen distance function that calculates the variation between a simulated value, and a value from the fixed dataset in terms of summary statistic $s_i$. The distances corresponding to each summary statistic are normalized to lie in $[0,1]$.
Let $M$ be the number of simulations used to accumulate $N$ accepted samples ${\bm \theta}_{acc}$, and therefore, the number of arm pulls. For a fixed value of $M$, the $K$ arms must be pulled in a sequence that maximizes the accumulated reward at the end of $M$ pulls. However, considering the ABC rejection sampling algorithm, the goal is to achieve $N$ accepted samples ${\bm \theta}_{acc}$ using as few simulations as possible.  
This also corresponds to selecting an arm such that the distance between simulated and observed data is minimized, thereby maximizing the reward.
Let $r_i^m$ be the reward for pulling the arm $s_i$ at the $m^{th}$ iteration. In order to maximize the accumulated reward, the quantity $R_E=E[\sum_{m=1}^M r_i^m]$ must be maximized, where $r_i^m \sim P_i$. Considering stochasticity, the expected total reward $R_E$ is maximized,
\begin{align}
R_E &= E\left[\sum_{m=1}^M r_i^m\right]  
= \sum_{m=1}^M E\left[r_i^m\right] 
= \sum_{m=1}^M \mu_{s^*_m},
\end{align}
where $s^*_m \in S$ for $m=1,2,..., M$, is the sequence of selected summary statistics or arm pulls. Maximizing the cumulative mean estimated reward corresponds to minimizing cumulative distance between ${\bf y}_{o}$ and ${\bf y}_{\bm \theta}$. The intuition is that this will lead to quick convergence of the ABC rejection sampler (in terms of number of simulations used), and will also improve the quality of parameter inference as each iteration will consist of selection of the most appropriate summary statistic for inference (corresponding to the least estimated distance).

Several strategies exist for solution of MAB problems. Popular strategies include $\epsilon$-greedy \cite{watkins1989learning} and its variants ($\epsilon$-first and $\epsilon$-decreasing), upper confidence bound (UCB) \cite{auer2002finite}, and Thomson sampling (TS) \cite{chapelle2011empirical}. For a detailed review of algorithms for solution of MABs, the reader is referred to \cite{kuleshov2014algorithms,zhou2015survey}. The $\epsilon$-first strategy \cite{vermorel2005multi} has been used for the purpose of experiments in this paper. It should be noted that the framework proposed herein is independent of any particular strategy, and in principle any algorithm that solves a MAB can be used as the $SelectStatistic$ method in Algorithm \ref{alg:rejectionStat}. The following text discusses the $\epsilon$-first strategy, and motivates its choice.

\subsection{The $\epsilon$-first strategy}
The $\epsilon$-greedy family of strategies is the most widely used and the simplest class of MAB algorithms. The $\epsilon$-first strategy is a variant of the $\epsilon$-greedy strategy that starts with a pure exploration phase where a random arm is chosen for the first $\epsilon$ fraction of pulls. Thereafter, a pure exploitation phase ensues where the arm corresponding to the highest mean estimated reward $\hat{\mu_{s_i^n}}$ is pulled. The value of $\epsilon$ is user-defined. It has been shown within a PAC framework that a total of $\mathcal{O}\Big(\frac{K}{\alpha^2} log \Big(\frac{K}{\delta}\Big)\Big)$ random arm pulls are sufficient to find an $\alpha$-optimal arm with probability at least ($1-\delta$) \cite{vermorel2005multi,even2002pac}. Empirical comparisons between various MAB strategies suggest that simpler algorithms like $\epsilon$-greedy outperform more sophisticated, theoretically sound approaches \cite{kuleshov2014algorithms,vermorel2005multi} on a majority of problems.

The intuition behind using the $\epsilon$-first strategy as a MAB solution lies in the fact that in the initial iterations of rejection sampling, the behavior of the various summary statistics is unknown. Therefore, it makes sense to perform pure exploration in order to gauge the efficacy and behavior of different statistics. As sampling proceeds, and the interplay of the distance function $d$ and various summary statistics is observed, and a more pragmatic approach can be followed in the form of exploitation. The formulation of exploration and exploitation in terms of the summary statistic selection problem is described below.

\subsection{Exploration and exploitation}
The classical formulation of exploration as per the $\epsilon$-first strategy is to select a summary statistic $s \in S$ at random for the first $\epsilon T$ iterations of the ABC rejection sampler, where $T$ is the total number of simulations or arm pulls allowed. The ABC rejection sampling problem setting typically does not include $T$, but instead includes the desired number of accepted samples $N$. It follows from the definition of $N$ and $T$, that for successful parameter inference, $N \leq T$. The fraction of exploration and exploitation iterations can also be defined in terms of $N$, i.e., explore for $\epsilon N$ iterations of rejection sampling, and exploit until $N$ samples have been accepted.

Exploitation typically corresponds to selecting an arm with the highest mean estimated reward.  The rewards are defined in terms of distances between the simulated and observed datasets calculated using the distance function $s$. Let $R$ be a $j \times K$ matrix storing the rewards calculated for each arm (summary statistic) pulled in all past iterations $j$. Let the $1 \times K$ vector $\bar{{\bf r}}$ denote the column-wise mean of $R$. Then $\bar{{\bf r}}$ can be used to represent the mean estimated reward of each arm in the current exploitation iteration, based on previous iterations. The arm with the highest estimated reward, $\max \bar{{\bf r}}$, is then chosen as the appropriate summary statistic $s$ for the current exploitation iteration. The statistic $s$ is then evaluated to compute distances, and rejection sampling proceeds according to Algorithm \ref{alg:rejectionStat}. The matrix $R$ is updated to include the reward corresponding to $s$.

\section{Experiments}
\label{sec:experiments}
The scalability of the proposed methodology is demonstrated on a challenging inference problem from molecular systems biology. We also compare and contrast it to two popular subset selection methods, namely approximate sufficiency (AS) and minimizing the entropy (ME). 

Discrete stochastic models of reaction networks based on the continuous-time Markov process formalism are frequently used to study gene regulatory networks \cite{Elowitz2002-ho}. These models are simulated with the stochastic simulation algorithm (SSA) due to Gillespie \cite{Gillespie1976-eq}, and the output of such simulations are statistically exact time series samples from the underlying Markov process model. Inference of the parameters of the models is highly challenging, since 
the stochastic nature of the simulator makes it computationally intractable to compute likelihoods for non-trivial models. As a consequence, ABC has gained popularity in this area \cite{Lillacci2013-yn,sunnaaker2013approximate,Lenive2016-zl}.  


A complex biochemical reaction network with oscillatory behavior \cite{vilar2002mechanisms} is used herein as a test problem. This network model involves $9$ chemical species undergoing $18$ chemical reactions, parameterized by $15$ reaction rate constants. The details of the chemical network such as its species and reaction definitions can be found in \cite{vilar2002mechanisms}. 
The model is implemented in StochSS \cite{drawert2016stochastic} and evaluated in Python using GillesPy \cite{abel2016gillespy}. All experiments involve the observed data ${\bf y}_o$ consisting of $300$ trajectories generated using GillesPy, with each trajectory spanning $1000$ time steps. The 15-dimensional search space of parameters during inference runs is represented by the following ranges, where parameter names are consistent with the notation in \cite{vilar2002mechanisms},
\begin{equation}
\label{eq:param_intervals}
\begin{aligned}
&\alpha_A \in [30,70],\quad 
&\alpha_A^* \in [200,600],\quad
&\alpha_R \in [0,1],\quad
&\alpha_R^* \in [30,70],\quad
&\beta_A \in [30,70],\\
&\beta_R \in [1, 10],\quad
&\delta_{MA} \in [1, 12],\quad
&\delta_{MR} \in [0,1],\quad
&\delta_A \in [0,2],\quad
&\delta_R \in [0,0.5],\\
&\gamma_A \in [0.5,1.5],\quad
&\gamma_R \in [0.5,1.5],\quad
&\gamma_C \in [1,3],\quad
&{\theta}_a \in [30,70],\quad
&{\theta}_r \in [80,120].
\end{aligned}
\end{equation}

The parameter values for the observed data were fixed as $\mathbf{\theta}=\{50,500,0.01,50,50,5,10,0.5,1,0.2,1,1,2,50,100\}$, i.e., the approximate center of the intervals \eqref{eq:param_intervals}. These parameter values give rise to reliable, but noisy oscillations \cite{vilar2002mechanisms}. Candidate pools of summary statistics are generated using the TSFRESH \cite{christ2016distributed} time series feature extraction framework. TSFRESH supports extraction of more than $700$ features, or summary statistics. The abctools \cite{nunes2015abctools} R library is used for evaluating AS and ME subset selection methods.

\subsection{Accuracy and scalability}
The inference problem detailed in the previous section is solved for summary statistic pools of sizes varying from $K=10$ to $K=200$ using the proposed method, as well as with traditional ABC with subset selection based on approximate sufficiency (AS) and minimizing the estimated entropy (ME) \cite{prangle2015summary} as a pre-processing step. The summary static pools were created by randomly selecting $K$ statistics to be generated by TSFRESH \cite{christ2016distributed} out of a possible total of 700. The random selection tests the flexibility of the approach, and increasing subset sizes test the scalability of the proposed methodology. The ABC acceptance threshold $\tau$ was set to accept samples with distances corresponding to top $5 \%$ of normalized distance values, i.e., $\tau=0.05$, where $\tau \in [0,1]$. The desired number of accepted samples $N$ were set to $100$ in order to get a reliable estimate of the mean of inferred parameters. The total number of allowed simulations $M$ is set to $300$, and $\epsilon$ in the MAB algorithm is set to $0.5$ to achieve equal balance between exploration and exploitation.

\begin{table}
\caption{Inference error for varying size of candidate summary statistics pool $K$ evaluated for approximate sufficiency (AS), minimizing the estimated entropy (ME) in the posterior and the proposed MAB approach. The reported values are mean absolute error (MAE) in inferred parameters with respect to actual parameters, repeated over 5 runs. Dashes indicate cases where code execution was unsuccessful due to out-of-memory exceptions.}
\label{tab:resultsMAE}
\centering
\scalebox{0.75}{
\begin{tabular}{lcccccccc}
\toprule
 & $K=10$ & $K=15$ & $K=20$ & $K=25$ & $K=30$ & $K=50$ & $K=100$ & $K=200$\\
\midrule
AS & ${\bf 7.77} \pm 0.98$ & $7.74 \pm 0.37$ & $7.65 \pm 0.99$ & $8.59 \pm 1.29$ & $-$ & $-$ & $-$ & $-$ \\
ME & $7.87\pm 0.98$ & ${\bf 7.72} \pm 0.97$ & $7.63 \pm 0.74$ & $-$ & $-$ & $-$ & $-$ & $-$ \\
MAB& $7.81 \pm 0.79$ & $8.18 \pm 0.95$ & ${\bf 7.04} \pm 0.26$ & ${\bf 8.21} \pm 1.00$ & $8.00 \pm 0.56$ & $7.47 \pm 0.62$ & $7.62 \pm 0.91$ & $8.07 \pm 0.53$ \\
\bottomrule
\end{tabular}}
\end{table}

Table \ref{tab:resultsMAE} shows the mean absolute error (MAE) in inferred parameters for varying size of summary statistics candidates pool ($K$). As can be seen, all summary statistic selection methods achieve similar inference quality across pool sizes. However, both AS and ME failed to select a subset for $K>25$ due to out-of-memory exceptions. The MAB approach, however, outperforms AS and ME in terms of efficiency and scalability.

\begin{table}
\caption{Total execution time for performing parameter inference, and computation time spent purely on summary statistic selection for varying size of candidate summary statistics pool $K$, evaluated for approximate sufficiency (AS), minimizing the estimated entropy (ME) in the posterior and the proposed MAB approach. The reported values are in seconds (s). Dashes indicate cases where code execution was unsuccessful due to out-of-memory exceptions. The values depict the mean and standard deviation over 5 runs.}
\label{tab:resultsTime}
\centering
\scalebox{0.63}{
\begin{tabular}{lccccccc}
\toprule
 & $K=10$ & $K=15$ & $K=20$ & $K=25$ & $K=30$ & $K=50$ & $K=200$\\
\midrule
\multicolumn{8}{l}{Total time taken for parameter inference:}\\
\midrule
AS & $2017.71 \pm 106.60$ & $1895.46 \pm 150.00$ & $2033.49 \pm 200.50$ & $2977.05 \pm 170.66$ & $-$ & $-$ & $-$ \\
ME & $1762.67 \pm 8.37$ & $1892.64 \pm 21.45$ & $6431.38 \pm 379.28$ & $-$ & $-$ & $-$ & $-$ \\
MAB& ${\bf 1161.53} \pm 90.09$ & ${\bf 1004.76} \pm 51.23$ & ${\bf 1044.11} \pm 1044.11$ & ${\bf 1034.48} \pm 34.10$ & $1019.28 \pm 89.79$ & $1064.36 \pm 35.34$ & $1057.26 \pm 153.22$ \\
\midrule
\multicolumn{8}{l}{Time spent purely on summary statistic selection:}\\
\midrule
AS & $0.18 \pm 0.06$ & $0.78 \pm 0.24$ & $13.50 \pm 1.03$ & $1100.52 \pm 33.23$ & $-$ & $-$ & $-$ \\ 
ME & $3.32 \pm 0.24$ & $128.21 \pm 4.93$ & $4576.26 \pm 191.95$ & $-$ & $-$ & $-$ & $-$ \\
MAB& ${\bf 0.09} \pm 0.01$ & ${\bf 0.10} \pm 0.01$ & ${\bf 0.02} \pm 0.00$ & ${\bf 0.13} \pm 0.02$ & $0.12 \pm 0.03$ & $0.17 \pm 0.03$ & $0.46 \pm 0.10$ \\
\bottomrule
\end{tabular}}
\end{table}
Table \ref{tab:resultsTime} shows the total time taken to solve the inference problem, as well as the time purely spent on summary statistics selection. As can be seen, the MAB approach outperforms both subset selection methods across the board. The AS approach is more efficient than ME, but the cost of both methods grow very rapidly as $K$ becomes large. In contrast, the MAB approach successfully handles candidate pools as large as $K=200$.

In order to validate the correctness of results obtained using the MAB-based summary statistic selection method, deeper analysis was performed on one of the sample runs, namely the case of $K=100$. It was observed that three summary statistics were ranked highly during the exploitation phase. Two of them were based on mass quantiles that calculate the proportion of mass concentration to the left of a certain part of the time series. The third summary statistic was based on Fourier coefficients of the one-dimensional discrete Fourier transform using the Fast Fourier Transform (FFT) algorithm. 

A separate ABC rejection sampling run was performed using only the highest ranked summary statistic. The run achieved the desired $N=100$ accepted samples after $183$ simulations, with MAE being $6.90$. Using the top-ranked statistic should intuitively yield faster convergence and lower MAE than the MAB-based approach, since exploration is avoided. The results confirmed this intuition. This also highlights the fact that while the MAB approach works as a black-box inference method with no need for pre-processing, it can also be useful for feature engineering for subsequent, optimized runs.  

Further, the top three summary statistics were used together in a separate run of ABC rejection sampling. The summary statistics were combined using the Euclidean norm ${\lVert S \rVert}_2$ of $S$. The run fulfilled acceptance criteria within only $124$ simulations, with MAE being $8.12$. The fast fulfillment points towards the informativeness of the top three features identified by the MAB-based approach. Contrarily, using three and ten randomly selected summary statistics for the purpose of comparison did not result in desired acceptance within the relatively large allowed budget of $1000$ simulations. This points towards the importance of well chosen summary statistics towards effective parameter inference, especially for complex and high-dimensional inference problems.

\section{Discussion and future work}
\label{sec:discussion}
The proposed approach was designed with the goal to meet the needs of highly challenging inference problems where hand-curated optimized summary statistics are not available \emph{a priori}, but where large pools of possible statistics are readily available. As explained in Sec. \ref{sec:abc}, existing methods either require extensive pre-processing (subset selection methods), require problem specific knowledge or training data (auxiliary likelihood methods), or lose the interpretability of the statistics (projection methods). There have also been recent efforts to develop non-parametric ABC algorithms that do away with the need to manually select summary statistics \cite{park2016k2}. However, summary statistics ease analysis and understanding of complex problems, such as the test problem in this work.
The MAB-based dynamic summary statistic selection method is designed keeping in view the considerations discussed above. Initial experiments in this paper have shown the proposed method to be highly promising. The proposed method was able to infer a relatively large number of parameters ($15$) in a complex non-linear model of a gene regulatory network, using a very small number of simulations. In contrast to the existing subset selection methods that were tested, the MAB approach scaled to pools with hundreds of statistics, although all methods managed to achieve high-quality inference for smaller pool sizes. The proposed method can also be used in more sophisticated ABC formulations like ABC-sequential Monte Carlo (ABC-SMC) \cite{beaumont2009adaptive}.  

A distinct advantage of the method is that it is completely black-box in nature and does not require any pre-processing. This makes it possible to use in a wide variety of applications with minimal modifications compared to standard ABC (selecting an appropriate value of $\epsilon$). It also makes large-scale automated summary statistic analysis possible, wherein the user does not have to hand-curate features to be tested. One can simply input hundreds of summary statistics, and only analyze in detail the ones used by the MAB solution during exploitation, for instance. As demonstrated here, this capability is very useful when high-throughput libraries for summary statistics, like TSFRESH, are available. The MAB-based method also maintains the interpretability of summary statistics. It merely acts as a filter to rapidly find the most ideal summary statistic for the inference problem. Moreover, the approach is computationally efficient and does not add substantial computational burden over and above rejection sampling.


Future work includes comparisons with highly-tuned ABC setups and methods in Sec. \ref{sec:abc} that make use of domain-specific knowledge.
Different formulations of exploration and exploitation will also be evaluated. Although the $\epsilon$-first strategy is used in this work, the proposed framework is independent of a particular class of MAB solutions, and in future other strategies will also be explored. 

A current characteristic of the proposed method is the fact that only one summary statistic is chosen from the candidates pool in each iteration. Although, during a complete run of Algorithm \ref{alg:rejectionStat}, a distribution over multiple statistics is considered.  It may be the case that a handful of summary statistics taken together offer more information to the distance function than the single best summary statistic in each iteration. Future work includes exploring this question in detail, both theoretically and empirically.

\section{Conclusion}
\label{sec:conclusion}
A novel methodology for performing multi-statistic parameter inference using the approximate Bayesian computation rejection sampling algorithm was presented in this paper. The problem of dynamically selecting the most appropriate summary statistic in each iteration of rejection sampling was formulated as a multi-armed bandit problem. The method does not require any prior problem-specific knowledge or pre-processing, and is highly scalable and computationally efficient. The efficacy of the proposed method was demonstrated by inferring parameters of a large scale stochastic biochemical reaction network. The proposed approach was shown to efficiently handle appropriate summary statistic selection from a pool of hundreds of candidate statistics.



\bibliographystyle{plainnat}
\bibliography{refs}







\end{document}